\documentclass{article}

 \usepackage[preprint]{neurips_2026}

\usepackage[utf8]{inputenc} 
\usepackage[T1]{fontenc}    
\usepackage{hyperref}       
\usepackage{url}            
\usepackage{booktabs}       
\usepackage{amsfonts}       
\usepackage{nicefrac}       
\usepackage{microtype}      
\usepackage[table]{xcolor}         

\usepackage{wrapfig}   
\usepackage{graphicx}

\usepackage{listings}
\usepackage{algorithm}
\usepackage{subcaption}

\usepackage{amsmath}

\newif\ifdraft
\drafttrue

\ifdraft
\newcommand{\noam}[1]{{\color{cyan}[\textbf{Noam:} #1]}}
\newcommand{\raanan}[1]{{\color{red}[\textbf{Raanan:} #1]}}
\newcommand{\dani}[1]{{\color{magenta}[\textbf{Dani:} #1]}}

\else
\newcommand{\noam}[1]{}
\newcommand{\raanan}[1]{}
\newcommand{\dani}[1]{}

\fi

\newcommand{\ie}{\emph{i.e.}}
\newcommand{\method}{\emph{CBS}}

\title{Complexity-Balanced Diffusion Splitting}

\author{
  Noam Issachar \quad
  Dani Lischinski \quad
  Raanan Fattal 
  \\ \\
  The Hebrew University of Jerusalem
}

\begin{document}

\maketitle

\newcommand{\fix}{\marginpar{FIX}}
\newcommand{\new}{\marginpar{NEW}}

\begin{abstract}
Standard continuous-time generative models rely on monolithic architectures that must navigate vastly different signal regimes, from isotropic noise to intricate data distributions. While scaling model capacity improves performance, deploying a massive network uniformly across the entire generative timeline is inherently inefficient. 
In this work, we propose \textit{Complexity-Balanced Splitting} (\method), a principled framework for temporal capacity allocation that distributes the generative workload across multiple specialized sub-networks. Grounded in function approximation theory and de Boor’s equidistribution principle, \method\ partitions the diffusion timeline into segments of equal approximation burden, allocating more representational capacity to regions where the generative dynamics are more difficult to model. 
To estimate this local complexity, we introduce two complementary and tractable monitor functions: a spatial measure based on the flow’s Dirichlet energy, and a geometric measure based on the acceleration of the sampling trajectories. Using a lightweight auxiliary model to estimate these complexity profiles, our approach eliminates the need for heuristic temporal splits or computationally expensive search procedures.
Extensive evaluation across multiple architectures (SiT, JiT, and UNet) and datasets demonstrates that \method\ consistently improves synthesis quality without increasing per-step inference cost. In particular, \method\ improves FID by \textasciitilde$35\%$ on SiT-XL with CFG relative to naive temporal partitioning. Project page is available at \url{https://noamissachar.github.io/CBS/}.
\end{abstract}

\section{Introduction}
\label{sec:intro}

Diffusion models~\citep{ho2020denoising,song2020score} have established themselves as a dominant paradigm for high-fidelity generative modeling. A primary driver of this success is their remarkable scalability, which offers a reliable pathway to translate increased model capacity and compute into state-of-the-art visual synthesis~\citep{flux2024}. Yet, standard diffusion frameworks rely on a \emph{monolithic} neural architecture to execute the entire denoising process. In this setting, a single model must operate across vastly different signal regimes, ranging from near-isotropic noise to highly structured data distributions, continuously adapting its function from coarse structural formation to fine-grained refinement.

To cope with this heterogeneity, the standard practice is to scale up the model~\citep{liang2024scaling, peebles2023scalable}. However, this strategy is inherently inefficient: the full scaled-up network is deployed uniformly across all timesteps, despite the fact that no individual denoising regime warrants such massive capacity on its own. As a more efficient alternative, one can distribute capacity temporally by training multiple specialized networks, each responsible for a different phase of the denoising process~\citep{balaji2022ediff, park2023denoising, feng2023ernie, park2024switch, lee2024multi}. 
Because only one sub-network is evaluated at each timestep, this design enables scaling the total parameter count without increasing per-step inference cost (FLOPs).

A central challenge in this paradigm is determining how to partition the diffusion timeline. Existing approaches typically rely on heuristic splits~\citep{feng2023ernie} or computationally expensive search procedures over candidate boundaries~\citep{balaji2022ediff}, which involve training multiple large-scale models under different partitioning schemes, most of which are ultimately discarded. Thus, these methods lack a \emph{principled criterion} for distributing capacity over time, limiting both their efficiency and generality.

In this work, we propose \textit{Complexity-Balanced Splitting} (\method), a principled framework for temporal capacity allocation in diffusion models. Our key idea is to consider the problem from a functional approximation standpoint and apply the de Boor equidistribution principle 
to partition the diffusion timeline into segments of equal approximation burden. Intuitively, this allocates more representational capacity to regions where the generative dynamics are more difficult to model, leading to a more uniformly accurate flow field and thus improved sample quality.

We consider two complementary monitor functions for estimating the local approximation burden of the target flow. The first is derived from neural approximation bounds based on the spatial variation of the flow field. While existing bounds are computationally intractable in high dimensions, we relate them to the flow's Dirichlet energy, which can be estimated efficiently. The second monitor function captures the geometric complexity of the sampling trajectories induced by the flow field through their second-order time derivative, which is an established complexity measure in geometric modeling.


We evaluate our approach across multiple architectures and datasets, demonstrating consistent improvements in synthesis quality without increasing per-step inference cost. Extensive evaluation of both monitor functions shows a substantial improvement over naive time-splitting and the ability to reach near-optimal results without a computationally prohibitive exhaustive search. Specifically, we show an improvement of \textasciitilde$15\%$ without CFG~\citep{ho2022classifier} to \textasciitilde$35\%$ with CFG in FID scores on SiT-XL with respect to naive splitting, demonstrating that complexity-based partitioning closely matches or exceeds the performance of more expensive search-based approaches. Furthermore, ablation studies confirm that aligning temporal boundaries with the local geometric complexity of its sampling trajectories leads to more balanced learning and improved robustness.

\section{Preliminaries}
\label{sec:background}

We provide here the theoretical background for our approach, specifically, we frame generative modeling as continuous-time velocity prediction and review the theory behind global function approximation and domain decomposition. We use the latter to derive our principled time-splitting scheme designed to evenly allocate the representational workload across the entire generative process.

\subsection{Diffusion Models via Velocity Prediction}
\label{subsec:velocity_prediction}
Recent advances unify diffusion models~\citep{ho2020denoising} and flow matching (FM)~\citep{lipman2022flow} under the framework of continuous-time interpolants. These methods construct a time-augmented state trajectory, where the intermediate state $x_t$ at continuous time $t \in [0, 1]$ smoothly bridges a tractable noise prior $x_0 \sim \mathcal{N}(0, I)$ and a target data sample $x_1 \sim q(x_1)$. Over the course of this trajectory, the generative process transitions through profoundly distinct phases, typically shifting from broad structural formation at high noise levels to high-frequency detail refinement near the data manifold. 

Whether formulated as flow matching~\citep{lipman2022flow} or $v$-prediction diffusion~\citep{salimans2022progressive}, modern methods train a neural network $v_\theta(x_t, t)$ to predict the trajectory's true instantaneous velocity, $u(x_t, t) = \frac{d}{dt}x_t$. The unified optimization objective is to regress the network against this ground-truth velocity,
\begin{equation}
    \mathcal{L} = \mathbb{E}_{t, x_0, x_1} \left[ \left\| v_\theta(x_t, t) - u(x_t, t) \right\|^2 \right].
    \label{eq:fm_loss}
\end{equation}
By viewing the generative process through this lens, the network's only task is to approximate the local velocity field of the trajectory.

\subsection{Global Function Approximation and Modeling Error}
\label{subsec:modeling_error}
Our central challenge is that a finite-capacity neural network must approximate a target velocity field whose complexity varies substantially over time. To motivate our time-splitting strategy, we draw on classical approximation theory and domain decomposition methods, which study how approximation error should be distributed across a target domain.

\paragraph{Domain Decomposition and Complexity Distribution.}
To circumvent the limitations of global approximation, a powerful strategy is the domain splitting that we follow. Rather than fitting a single global model to $u$ over the entire domain $\Omega$, it is partitioned into a set of $N$ disjoint intervals. Let $\Omega = [0, 1]$ be partitioned by a set of nodes (or knots):
\begin{equation}
 0 = t_0 < t_1 < t_2 < \dots < t_{N-1} < t_N = 1, 
 \label{eq:time_splits}
\end{equation}
on each interval $\Omega_i = [t_{i-1}, t_i]$, we deploy a separate localized model $\hat{f}_i$. Assuming all the models are of the same form and posses the same modeling capacity, we transform the problem to optimally allocating sub-problems, \ie, modeling intervals, of equal complexity. The fundamental challenge then becomes: \textit{How should we choose the splitting points $\{t_i\}$ such that the total modeling complexity is optimally distributed across the domain?}

\paragraph{The De Boor Principle and Equidistribution.}
The theoretical framework for optimal domain partitioning originates in approximation theory, specifically in the study of spline approximations formulated by de Boor~\citep{deboor1973good}. The core concept is that the nodes should be densely clustered in regions where the target function $f$ is highly complex, and sparsely distributed in regions where $f$ is smooth.

To formalize this idea, a \emph{monitor function} $m(t) > 0$ is introduced to quantify the local approximation burden of the target function $f$ at time $t\in [0, 1]$. Thus, regions where $m(t)$ is large requiring finer partitioning, while regions with small $m(t)$ can be modeled using wider intervals. In classical polynomial spline approximation of degree $k$, the monitor function is often chosen proportional to a fractional power of the $(k+1)$-th derivative of $f$~\cite{curvature}. 

The modeling error at the $i$-th interval $[t_{i-1}, t_i]$ can be bounded by integrating the monitor function over that interval,
\begin{equation}
E_i \approx C \int_{t_{i-1}}^{t_i} m(t) \, dt \end{equation}
where $C$ is a constant independent of the partition.

The objective of minimizing the maximal approximation error across all sub-intervals, leads to a min-max optimization problem for the knot placement:
\begin{equation}
\min_{\{t_i\}} \max_{1 \le i \le N} \int_{t_{i-1}}^{t_i} m(t) \, dt 
\end{equation} 
The De Boor Principle \citep{deboor1973good} shows that this problem is solved when the error bound is \textit{equidistributed} across all intervals. In other words, the optimal partition is achieved when the integral of the monitor function is strictly equal for every sub-interval, yielding the final equidistribution,
\begin{equation}
    \int_{t_{i-1}}^{t_i} m(t) \, dt = \frac{1}{N} \int_{0}^{1} m(t) \, dt \quad \text{for all } i = 1, 2, \dots, N 
\end{equation}
As a result, each local model faces a comparable representational burden which aligns well with having the same modeling capacity. This principle forms the foundation of our approach to temporal capacity allocation in diffusion models.

\begin{wrapfigure}[7]{r}{0.3\linewidth}
    \vspace*{-1\topskip}
    \centering
    \includegraphics[width=\linewidth]{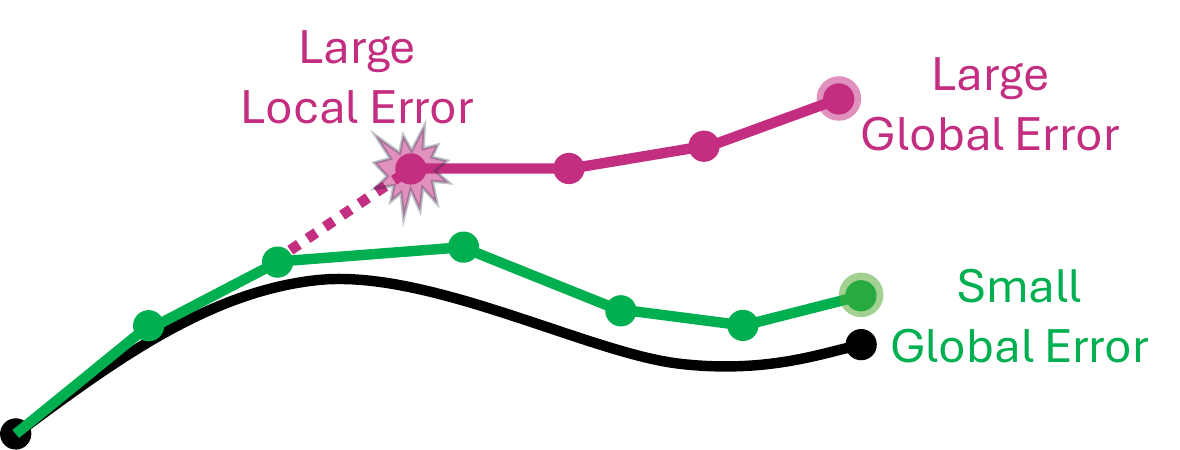}
    \caption{Maximal local errors govern ODE approximation.}
    \label{fig:ode_error}
\end{wrapfigure}
 Finally, let us consider our specific context where the target function $u$ is a flow field which is path-integrated during sampling generation. In this case, once a large error is encountered, the resulting path diverges from its true course and the misalignment is maintained forward in time, even if the flow is fairly accurate in the subsequent steps, as illustrated in Fig.~\ref{fig:ode_error}. Thus, the objective of minimizing the maximal error bound is specifically directed toward better sampling quality~\citep{gronwall1919note}. 

In this respect we note that standard training paradigms for denoising models, such as DDPM~\citep{ho2020denoising} and flow matching~\citep{lipman2022flow}, optimize for the expected error averaged across the entire denoising interval (as seen in Eq.~\ref{eq:fm_loss}). This approach fundamentally diverges from this theoretically motivated strategy for minimizing the maximum instantaneous error.

\section{Method}
\label{sec:method}

To derive our time-splitting scheme based on these principles, we must define a meaningful monitor function $m(t)$. We propose two such functions: the first assesses the target flow field's complexity, via Barron's error bound~\citep{barron2002universal} (Sec.~\ref{subsec:dirichlet_barron_work}), while the second evaluates the geometric complexity of the resulting flow paths (sampling trajectories) via classical curve approximation measures (Sec.~\ref{subsec:curvature_work}). Together, these two approaches encompass the primary methods for quantifying the analytical complexity of a generative flow field. Finally, in Sec.~\ref{subsec:training_inference} we describe our training and inference pipelines.

\subsection{Modeling Burden via Dirichlet Spectral Energy}
\label{subsec:dirichlet_barron_work}

Barron's theorem~\citep{barron2002universal} relates the approximation error of a network to the spectral properties of the target function. Formally, for a target function $f$ defined on a bounded domain of radius $r$, and data drawn from a probability distribution $p$, a feedforward network $f_n$ with $n$ parameters achieves an expected Mean Squared Error (MSE) bounded by:
\begin{equation}
\label{eq:barron}
    \varepsilon_{f_n} = \mathbb{E}_{p_t} \left[ \|f(x) - f_n(x)\|^2 \right] \le \frac{4 r^2 C_f^2}{n},
\end{equation}
where $C_f$ is the spectral complexity of the target function, formally defined as the first absolute moment of the function's Fourier transform: 
\begin{equation}
    C_f = \int_{\mathbb{R}^d} \|\omega\| |\hat{f}(\omega)| d\omega.
\end{equation}
A smooth, gently changing function possesses a low $C_f$, whereas a highly non-linear function exhibiting sharp transitions is characterized by a higher $C_f$.

In our context of modeling denoising flows, the target function is the instantaneous vector field, $v_t(x)$, discussed in Sec.~\ref{subsec:velocity_prediction}. Consequently, we cannot assess $C_{v_t}$ in practice due to the flow's exceedingly high spatial dimension. Instead, we obtain a formal bound on $C_{v_t}$ based on the vector field's global spatial variation, measured by its \emph{Dirichlet energy}
\begin{equation}
\label{eq:dirichlet}
    E_D(v_t) = \frac{1}{2} \int_{\mathbb{R}^d} \|\nabla_x v_t(x)\|^2 dx.
\end{equation}
Intuitively, rapid spatial variation across the domain is associated with high spectral energy. By applying Parseval's identity for gradients of functions~\citep{rudin2021principles}, the Dirichlet energy translates directly to the frequency domain:
\begin{equation}
    E_D(v_t) = \frac{1}{2(2\pi)^d} \int_{\mathbb{R}^d} \|\omega\|^2 \|\hat{v}_t(\omega)\|^2 d\omega.
\end{equation}

Finally, by applying the Cauchy-Schwarz inequality over an effective frequency bandwidth, we can bound the spectral complexity $C_{v_t}^2$ using the Dirichlet energy:
\begin{equation}
\label{eq:cs_bound}
    C_{v_t}^2 \le K \cdot E_D(v_t),
\end{equation}
where $K$ is a domain-dependent constant relating to the support of the field's frequency spectrum. Substituting this bound back into Barron's theorem establishes a direct, computable monitor function based on the global spatial roughness of the flow and the expected approximation error for a parameter budget $n$:
\begin{equation}
\label{eq:bound}
    \varepsilon_{f_n}  \le K'  \frac{E_D(v_t)}{n} = m(t).
\end{equation}
The full derivation of this bound, including the formulation of the bandwidth constant $K$, is provided in Appendix~\ref{sec:appendix_a}.

\subsection{Modeling Burden via Path Acceleration}
\label{subsec:curvature_work}
While the Dirichlet energy bounds the modeling burden from the perspective of the spatial flow field complexity, we explore an alternative and complementary bound that we derive by analyzing the \textit{temporal complexity} of the underlying sampling trajectories. 

For any given initial noise state $x_0$, the flow field $u$ defines a continuous trajectory $x_t$ in $\mathbb{R}^d$, as a function of time $t \in [0, 1]$. The modeled flow $v_\theta$ gives rise to an approximate trajectory $\hat{x}_t$, thus modeling error between the two naturally arise. The approximation theory literature, ranging from classic $n$-degree polynomials or Fourier series to recent specific classes of deep neural networks~\citep{YAROTSKY2017103}, encounters a general bound on the global approximation error on curves, given by
\begin{equation}
    \|x_t - \hat{x}_t\|_\infty \le C \frac{L_k}{n^k}
    \label{eq:curve_bound}
\end{equation}
where $n$ represents the model capacity (parameter count), $k$ denotes the order of the derivative used to assess the trajectory's smoothness, and the constant $C$ is independent of $n$ and the trajectory. In this global bound, the curve's $k$-th derivative term $L_k$ is maximized over the entire interval, by $L_k = \sup_t \left\| d^k x_t/dt^k \right\|$, and is therefore overly pessimistic as it may vary substantially over time. 

To overcome this pessimistic global assessment, we employ the magnitude of the $k$-th derivative as a timewise monitor function, integrating it over the partitioned time segments to achieve the optimal, equidistributed splitting.

While this bound suggests that higher-order smoothness (larger $k$) guarantees faster asymptotic convergence, the ability to assess $L_k$ reliably diminishes in approximate practical settings. The choice of $k=1$ bounds the error based on the path's maximum velocity. This confounds geometric complexity with traversal speed. A flow field scaled by a constant factor traverses the same geometric path faster; while this does not inherently increase the complexity of the function the network must learn (it merely scales the output). Moreover, it penalizes for spatial displacement rather than the actual non-linearity of the trajectory. 

Instead, as a monitor function we opt for $k=2$, which bounds the error using the path's acceleration, 
\begin{equation}
    m(t) = \left\| \frac{d^2 x_t}{dt^2} \right\|,
\end{equation}
which effectively filters out the constant-velocity displacement and isolates the curviness of the path. In case of trajectories with relatively constant velocity magnitudes, the second-order derivative effectively approximates the curvature, a well-established measure of complexity in classical geometric modeling \citep{curvature}.

\subsection{Deriving the Time-Splitting Scheme}

Both monitor functions $m(t)$ discussed above require access to the target flow field and its induced trajectories. To obtain these, we first train a single auxiliary network over the entire time frame $t \in [0,1]$. Because this network is used solely for estimating the temporal boundaries $t_i$ from Eq.~\ref{eq:time_splits}, a highly approximate model is sufficient. By training a smaller architecture on a fraction of the dataset (10\% in our implementation) for fewer epochs, we obtain a sufficiently reliable time-splitting with minimal overhead, as validated in Sec.~\ref{subsec:boundary_efficiency}.

Using this auxiliary flow network, we generate $K$ sampling trajectories, $\{x_t^k\}_{k=1}^K$, which serve for estimating the monitor functions $m(t)$ over a uniform temporal grid (100 grid points in our implementation).

\paragraph{Dirichlet Energy Monitor Function.} Evaluating $E_D(v_t)$ is computationally demanding because the spatial gradient term $\|\nabla_x v_t(x)\|^2$ represents the squared Frobenius norm of the $d \times d$ Jacobian matrix, and $d$ is exceedingly large in practice. To bypass materializing the full Jacobian, we employ randomized trace estimators~\citep{hutchinson1990stochastic} that rely on efficient Jacobian-Vector Products (JVPs). At each grid point $t$, our $K$ trajectories provide samples from the marginal distribution $p_t$. We approximate the spatial integral in Eq.~\ref{eq:dirichlet} by averaging the JVP-based gradient estimates across these $K$ points.

\paragraph{Path Acceleration Monitor Function.} The $K$ trajectories evaluated on a discrete temporal grid. We approximate the second-order time derivative via a first-order finite difference over the velocity field $v_t$. Specifically:
\begin{equation}
    m(t) = \frac{1}{K}\sum_{k=1}^K \|v_{t+\Delta t}(x_{t+\Delta t}^k) - v_{t}(x_{t}^k)\|,
\end{equation}
which follows from the relation $dx_t/dt = v_t(x_t)$.

Finally, given $m(t)$ evaluated across the uniform temporal grid, we approximate de Boor's equidistribution principle discretely to determine the final time partition points, $t_i$. Specifically, we compute the cumulative sum of $m(t)$ and select the boundaries as the grid points that partition the total accumulated monitor
value equally among the $N$ segments. 

Formally, the $i$-th split $t_i$ is chosen as the grid point where the cumulative sum most closely approximates $\frac{i}{N} \sum_j m(t_j)$.

\definecolor{codeblue}{rgb}{0.25,0.5,0.5}
\definecolor{codekw}{rgb}{0.85, 0.18, 0.50}

\definecolor{codesign}{RGB}{0, 0, 255}
\definecolor{codefunc}{rgb}{0.85, 0.18, 0.50}

\lstdefinelanguage{PythonFuncColor}{
  language=Python,
  keywordstyle=\color{blue}\bfseries,
  commentstyle=\color{codeblue},  
  stringstyle=\color{orange},
  showstringspaces=false,
  basicstyle=\ttfamily\small,
  literate=
    {*}{{\color{codesign}* }}{1}
    {-}{{\color{codesign}- }}{1}
    {+}{{\color{codesign}+ }}{1}
    {sample_model_idx}{{\color{codefunc}sample\_model\_idx}}{1}
    {sample_t}{{\color{codefunc}sample\_t}}{1}
    {randn}{{\color{codefunc}randn}}{1}
    {randn_like}{{\color{codefunc}randn\_like}}{1}
    {find_active_model}{{\color{codefunc}find\_active\_model}}{1}
    {linspace}{{\color{codefunc}linspace}}{1}
    {metric}{{\color{codefunc}metric}}{1}
}

\lstset{
  language=PythonFuncColor,
  backgroundcolor=\color{white},
  basicstyle=\fontsize{9pt}{9.9pt}\ttfamily\selectfont,
  columns=fullflexible,
  breaklines=true,
  captionpos=b,
}

\begin{wrapfigure}[11]{r}{0.47\linewidth}
\vspace{-2.5em} 
\centering
\begin{minipage}{0.95\linewidth}

\begin{algorithm}[H]
\caption{CBS Training}
\label{alg:cbp_train}
\begin{lstlisting}
# models: N networks, bounds: N+1 splits
# x_1: data batch

i = sample_model_idx(N)
t = sample_t(bounds[i], bounds[i+1])

x_0 = randn_like(x_1)
x_t = t * x_1 + (1 - t) * x_0

# Predict velocity and compute loss
pred_v = models[i](x_t, t)
loss = metric(pred_v - (x_1 - x_0))
\end{lstlisting}
\end{algorithm}

\end{minipage}
\end{wrapfigure}

\subsection{Training and Inference}
\label{subsec:training_inference}
The resulting $t_i$ are used as the boundaries in both training and inference of \method\ as elaborated next.

\paragraph{Training.} During the optimization phase, each specialized network $v_{\theta_i}$ is trained independently using the standard velocity prediction objective (Eq.~\ref{eq:fm_loss}). However, its time domain is strictly bounded: the timestep $t$ is sampled exclusively from its designated interval, as summarized in Alg.~\ref{alg:cbp_train}. 
\paragraph{Inference.} Generating a novel sample is done by switching between the networks, using each one $v_{\theta_i}$ over its designated segment $[t_i,t_{i+1}]$. 

\section{Experiments}
\label{sec:experiments}
We implemented and evaluated \method\ to assess its performance, robustness and generality. As it is derived over mathematical bounds, our evaluation particularly aims to assess its theoretical tightness and demonstrate its practical superiority across multiple architectures, datasets, and partitioning size. We start by detailing our experimental settings across three generative domains (Sec.~\ref{subsec:setup}). We then present our time-splitting performance in terms of added generative accuracy as well as its scaling analysis (Sec.~\ref{subsec:main_results}). This is followed by validating the core premise of our approach in its ability to achieve close to optimal splitting (Sec.~\ref{subsec:optimal_splits}), and a direct comparison between our proposed monitor functions (Sec.~\ref{subsec:monitor_comparison}). Finally, we address the practical overhead of our method, demonstrating that these temporal boundaries can be estimated with negligible computational cost (Sec.~\ref{subsec:boundary_efficiency}).
\subsection{Experimental Setup}
\label{subsec:setup}

To rigorously evaluate \method, we design our empirical study across three distinct generative environments, each posing unique spatial and spectral challenges.
\paragraph{High-Fidelity Latent Synthesis (ImageNet-256).}
Our primary testbed is the ImageNet dataset at $256 \times 256$ resolution, which is a standard benchmark for complex, conditional image generation containing 1000 distinct classes. Modeling flows at this resolution is computationally prohibitive, hence we operate in the latent space of a pre-trained autoencoder. We utilize the Scalable Interpolant Transformer (SiT)~\citep{ma2024sit} as our baseline flow model architecture. SiT operates on latent patches and scales predictably, allowing us to cleanly evaluate our network ensemble against monolithic baselines of varying capacities (SiT-S, SiT-B, and SiT-XL).

\paragraph{Pixel-Space Synthesis (ImageNet-64).}
Operating in latent space often smooths out certain high-frequency features. To assess how well our method handles raw, uncompressed spatial gradients, we evaluate it also over pixel-space generation on the ImageNet at $64 \times 64$ resolution. We use the Just Image Transformer (JiT)~\citep{li2025back} as the flow model.

\paragraph{Unconditional Generation (CIFAR-10).}
Finally, we evaluate our method on the CIFAR-10 dataset ($32 \times 32$ resolution) in a purely unconditional setting. For this task, we swap the transformer backbone for a standard convolutional UNet architecture~\citep{ronneberger2015u}, demonstrating that the benefits of complexity-based partitioning are architecture-agnostic.

\paragraph{Evaluation Metrics.}
To quantitatively assess the performance of \method\ against the baselines, we employ a comprehensive suite of standard generative metrics. Our primary metric is the Fr\'echet Inception Distance (FID)~\citep{heusel2017gans}, which provides a holistic measure of both image fidelity and distributional diversity. We also report the Inception Score (IS)~\citep{salimans2016improved} as a secondary measure for class distinguishability and intra-class diversity. Finally, to disentangle the trade-off between synthesis quality and mode coverage, we report Precision and Recall~\citep{kynkaanniemi2019improved}.

\paragraph{Implementation Details.}
Unless otherwise specified, all experiments use $N=3$ specialized networks, and the path acceleration based monitor function $m(t)$. All monolithic baselines and our sub-networks are trained using the same default hyperparameters to ensure a fair comparison. The cumulative monitor energy used to derive our time-splits is pre-computed over a grid of 100 points, as described in Sec.~\ref{subsec:boundary_efficiency}. The exact training hyperparameters, sampling configurations, and further hardware specifics are detailed in Appendix~\ref{sec:appendix_impl}.

\subsection{Generative Performance and Network Scaling}
\label{subsec:main_results}

\renewcommand{\arraystretch}{1}
\begin{table*}[t!]

    \centering
    \footnotesize
    \setlength{\tabcolsep}{4pt}
    {
\caption{
    \textbf{Quantitative evaluation of SiT on ImageNet-256.} We compare the standard monolithic baseline, a naive uniformly partitioned ensemble (partitioned at $0.33, 0.66$), and our approach (\method) using complexity-derived boundaries ($0.4, 0.77$). Across all model capacities (S/2, B/2, and XL/2), our method yields consistent and significant improvements in FID, Inception Score (IS), Recall and Precision, both with and without Classifier-Free Guidance (CFG). Crucially, \method\ achieves these generative gains while strictly maintaining the exact same active parameter count and per-step inference cost (GFLOPs) as the standard monolithic baseline.
    }
    
    \label{tab:sit_metrics_scales}
    
    \begin{tabular}{@{}l|cc|cccc|cccc@{}}
    \toprule
        & & & \multicolumn{4}{c|}{\textbf{Without CFG}} & \multicolumn{4}{c@{}}{\textbf{With CFG}} \\
        \cmidrule(lr){4-7} \cmidrule(l){8-11}
        \textbf{Model} 
        & \textbf{\# Act. Par.} & \textbf{GFLOPs}
        & \textbf{FID$\downarrow$} & \textbf{IS$\uparrow$} & \textbf{Prec.$\uparrow$} & \textbf{Rec.$\uparrow$} 
        & \textbf{FID$\downarrow$} & \textbf{IS$\uparrow$} & \textbf{Prec.$\uparrow$} & \textbf{Rec.$\uparrow$} \\
    \midrule
        SiT-S/2 & 33 & 6.06 & 58.97 & 23.34 & 0.40 & 0.59 & 30.10 & 49.82 & 0.57 & 0.50 \\
        SiT-S/2 (0.33 0.66) & 33 & 6.06 & 58.82 & 23.23 & 0.40 & 0.59 & 29.83 & 52.17 & 0.57 & \textbf{0.51} \\
        \rowcolor{blue!10} \textbf{SiT-S/2 Ours} (0.4 0.77) & 33 & 6.06 & \textbf{50.87} & \textbf{29.57} & \textbf{0.44} & \textbf{0.62} & \textbf{18.61} & \textbf{72.73} & \textbf{0.62} & \textbf{0.51} \\
    \midrule
        SiT-B/2 & 130 & 23.01 & 34.84 & 41.53 & 0.52 & 0.64 & 16.79 & 84.75 & 0.66 & 0.55 \\
        SiT-B/2 (0.33 0.66) & 130 & 23.01 & 34.52 & 41.45 & 0.52 & 0.64 & 16.51 & 87.21 & 0.66 & \textbf{0.56}\\
        \rowcolor{blue!10} \textbf{SiT-B/2 Ours} (0.4 0.77) & 130 & 23.01 & \textbf{30.51} & \textbf{52.20} & \textbf{0.55} & \textbf{0.65} & \textbf{10.72} & \textbf{121.09} & \textbf{0.72} & \textbf{0.56} \\
    \midrule
        SiT-XL/2 & 675 & 118.64 & 18.04 & 73.90 & 0.63 & 0.64 & 6.53 & 162.23 & 0.73 & \textbf{0.56} \\
        SiT-XL/2 (0.33 0.66) & 675 & 118.64 & 17.97 & 73.85 & 0.63 & 0.64 & 6.24 & 165.29 & 0.74 & \textbf{0.56} \\
        \rowcolor{blue!10} \textbf{SiT-XL/2 Ours} (0.4 0.77) & 675 & 118.64 & \textbf{15.81} & \textbf{86.43} & \textbf{0.64} & \textbf{0.66} & \textbf{4.03} & \textbf{195.72} & \textbf{0.80} & \textbf{0.56} \\
    \bottomrule
    \end{tabular}}
\end{table*}
\renewcommand{\arraystretch}{1}

\renewcommand{\arraystretch}{1}
\begin{table*}[t!]
    \centering
    \footnotesize
    \setlength{\tabcolsep}{4pt}
    {
    \caption{
    \textbf{Quantitative evaluation of JiT-B/4 on ImageNet-64.} In order to verify that \method\ generalizes effectively to raw, pixel-space spatial gradients, we compare our complexity-based partitioning against the standard monolithic JiT-B/4 architecture and a uniform temporal split. All configurations maintain an identical inference cost of 131M activated parameters and 25 per-step GFLOPs. Even with this strict compute budget, our method achieves significant improvements in synthesis quality.
    }
    \label{tab:jit_imagenet64}
    
    \begin{tabular}{@{}l|cccc|cccc@{}}
    \toprule
        & \multicolumn{4}{c|}{\textbf{Without CFG}} & \multicolumn{4}{c@{}}{\textbf{With CFG}} \\
        \cmidrule(lr){2-5} \cmidrule(l){6-9}
        \textbf{Model} 
        & \textbf{FID$\downarrow$} & \textbf{IS$\uparrow$} & \textbf{Prec.$\uparrow$} & \textbf{Rec.$\uparrow$} 
        & \textbf{FID$\downarrow$} & \textbf{IS$\uparrow$} & \textbf{Prec.$\uparrow$} & \textbf{Rec.$\uparrow$} \\
    \midrule
        JiT-B/4 & 17.43 & 25.92 & 0.55 & 0.59 & 16.41 & 158.52 & 0.87 & 0.28 \\
        JiT-B/4 (0.33 0.66) & 16.39 & 28.04 & 0.57 & \textbf{0.61} & 15.08 & 179.39 & 0.88 & 0.28 \\
        \rowcolor{blue!10} \textbf{JiT-B/4 Ours} (0.1 0.38) & \textbf{15.02} & \textbf{29.86} & \textbf{0.59} & \textbf{0.61} & \textbf{13.93} & \textbf{199.38} & \textbf{0.89} & \textbf{0.30} \\
    \bottomrule
    \end{tabular}}
\end{table*}
\renewcommand{\arraystretch}{1}

\paragraph{Latent and Pixel-Space Synthesis.} 
To evaluate the core efficacy of \method, we compare our multi-network configuration against standard monolithic models. As demonstrated in Tab.~\ref{tab:sit_metrics_scales}, splitting the generative workload according to our complexity monitor functions on the latent ImageNet-256 allows our partitioned SiT architecture to achieve significantly superior synthesis quality (measured via FID and IS) without inflating the per-step inference FLOPs. Furthermore, we observe that these architectural benefits compound significantly when utilizing Classifier-Free Guidance (CFG), as the localized capacity helps resolve the complex spatial gradients introduced by the guidance term.

\renewcommand{\arraystretch}{1}
\begin{wraptable}[7]{r}{0.27\textwidth}
    \centering
    \footnotesize
    \setlength{\tabcolsep}{4pt} \caption{\textbf{Unconditional generation on CIFAR-10.}
    }
    \vspace{-0.2cm}
    \label{tab:cifar10_results}
    \begin{tabular}{@{}l|c@{}}
    \toprule
        \textbf{Model} 
        & \textbf{FID$\downarrow$} \\
    \midrule
        Baseline & 3.55 \\
        (0.33 0.66) & 3.52 \\
        \rowcolor{blue!10} \textbf{Ours} (0.2, 0.78) & \textbf{2.72} \\
    \bottomrule
    \end{tabular}
    \vspace{-10pt} 
\end{wraptable}
\renewcommand{\arraystretch}{1}

This performance generalizes beyond flows in latent-space. As shown in Tab.~\ref{tab:jit_imagenet64}, \method\ successfully isolates and resolves the raw, high-frequency spatial gradients in pixel-space ImageNet-64 using the JiT architecture, achieving significant improvements over the monolithic baseline. Finally, our UNet-based evaluation on unconditional CIFAR-10 (Tab.~\ref{tab:cifar10_results}) confirms that the method scales down effectively to smaller datasets and non-transformer architectures.

\renewcommand{\arraystretch}{1}
\begin{wraptable}[12]{r}{0.4\textwidth}
    \vspace*{-1\topskip}
    \centering
    \footnotesize
    \setlength{\tabcolsep}{4pt}
    {
    \caption{
    \textbf{Scaling \method\ across multiple networks.} We evaluate SiT-B/2 partitioned into $N$ networks using our complexity-derived boundaries.
    }
    \label{tab:model_scaling}
    
    \begin{tabular}{@{}l|cc|cc@{}}
    \toprule
        & \multicolumn{2}{c|}{\textbf{Without CFG}} & \multicolumn{2}{c@{}}{\textbf{With CFG}} \\
        \cmidrule(lr){2-3} \cmidrule(l){4-5}
        \textbf{$N$ Models} 
        & \textbf{FID$\downarrow$} & \textbf{IS$\uparrow$} 
        & \textbf{FID$\downarrow$} & \textbf{IS$\uparrow$} \\
    \midrule
        $N=1$ & 34.84 & 41.53 & 16.79 & 84.75 \\
        $N=2$ & 32.19 & 47.82 & 12.98 & 109.42 \\
        $N=3$ & 30.51 & 52.20 & 10.72 & 121.09 \\
        $N=4$ & 29.33 & 55.03 & 9.42 & 129.49 \\
    \bottomrule
    \end{tabular}}
\end{wraptable}
\renewcommand{\arraystretch}{1}

\paragraph{Scaling the Number of Networks ($N$).}
To demonstrate the scalability of \method, we evaluate its performance as the generative timeline is partitioned across an increasing number of specialized networks. Tab.~\ref{tab:model_scaling} presents the synthesis quality when using $N \in \{1, 2, 3, 4\}$ models. Rather than relying on heuristic splits that become exponentially harder to tune for a larger number of networks, our complexity-based metric seamlessly derives optimal boundaries for any arbitrary $N$. As shown, increasing the number of networks consistently improves both FID and Inception Score. This directly validates that progressively relieving localized capacity bottlenecks through finer, mathematically principled temporal partitioning yields reliable generative gains. While we use $N=3$ as our default configuration to balance synthesis quality with total training overhead, the continued scaling up to $N=4$ highlights the robustness and generality of our splitting criterion.

\renewcommand{\arraystretch}{1}
\begin{wraptable}[10]{r}{0.5\textwidth}
    \vspace{-15pt} 
    \centering
    \footnotesize
    \setlength{\tabcolsep}{4pt}
    \caption{
    \textbf{Empirical validation of boundary optimality on SiT-B/2.} Table reports the results obtained by perturbing our derived splits ($0.4, 0.77$).
    }
    \label{tab:boundary_ablation}
    \begin{tabular}{@{}l|cccc@{}}
    \toprule
        \textbf{Model} 
        & \textbf{FID$\downarrow$} & \textbf{IS$\uparrow$} & \textbf{Prec.$\uparrow$} & \textbf{Rec.$\uparrow$} \\
    \midrule
        
        SiT-B/2 (0.35 0.77) & 31.44 & 49.83 & 0.54 & \textbf{0.65} \\
        SiT-B/2 (0.45 0.77) & 31.21 & 50.11 & 0.54 & 0.64 \\
        SiT-B/2 (0.4 0.72) & 30.89 & 51.17 & 0.54 & 0.64 \\
        SiT-B/2 (0.4 0.82) & 31.54 & 50.13 & 0.54 & 0.64 \\
        \rowcolor{blue!10} \textbf{SiT-B/2 Ours (0.4 0.77)} & \textbf{30.51} & \textbf{52.20} & \textbf{0.55} & \textbf{0.65} \\
    \bottomrule
    \end{tabular}
\end{wraptable}
\renewcommand{\arraystretch}{1}

\subsection{Empirical Optimality of Complexity Boundaries}
\label{subsec:optimal_splits}

While our partitioning scheme is strictly derived based on theoretical error bounds as discussed in Sec.~\ref{sec:method}, we measure how close these specific splittings are to an optimal solution. This  is done by training multiple sets of networks over perturbed versions of our temporal splits (changing each boundary separately).

As shown in Tab.~\ref{tab:boundary_ablation}, matching the temporal splits exactly to intervals of equal cumulative complexity (our time-splitting) consistently yields the lowest FID. These results confirm the relevance of our monitor function as an accurate proxy for the empirical learning burden.

\renewcommand{\arraystretch}{1}
\begin{wraptable}[12]{r}{0.55\textwidth}
    \vspace{-10pt}
    \centering
    \footnotesize
    \setlength{\tabcolsep}{4pt}
    \caption{
    \textbf{Comparison of Monitor Functions.} Both monitor functions result in near-optimal solution in SiT-B/2, with improved scores for the path acceleration based function. 
    }
    \label{tab:monitor_comparison}
    \begin{tabular}{@{}l|cccc@{}}
    \toprule
        \textbf{Model} 
        & \textbf{FID$\downarrow$} & \textbf{IS$\uparrow$} & \textbf{Precision$\uparrow$} & \textbf{Recall$\uparrow$} \\
    \midrule
        
        \textbf{SiT-B/2 (Dirichlet)} & 31.25 & 49.95 & 0.54 & 0.64 \\
        \textbf{SiT-B/2 (Acceleration)} & \textbf{30.51} & \textbf{52.20} & \textbf{0.55} & \textbf{0.65} \\
        
    \midrule
        
        \textbf{JiT-B/2 (Dirichlet)}
        & 15.40 & \textbf{29.90} &  0.58 & 0.59\\
        \textbf{JiT-B/2 (Acceleration)} & \textbf{15.02} & 29.86 & \textbf{0.59} & \textbf{0.61} \\
        
    \bottomrule
    \end{tabular}
\end{wraptable}
\renewcommand{\arraystretch}{1}
\subsection{Comparison of Monitor Functions}
\label{subsec:monitor_comparison}

We compare here the accuracy achieved by the two monitor functions suggested in Sec.~\ref{sec:method}: the spatial Dirichlet energy and the temporal path acceleration. Tab.~\ref{tab:monitor_comparison} presents the performance of both SiT-B/2 and JiT-B/2 when partitioned using each monitor function. While both metrics appear close to the optimal solution on SiT-B/2, the path acceleration-based monitor obtains better FID scores. For this reason, we use it as the default monitor in the rest of our tests. We attribute its greater success to the fact that it measures the final sampling accuracy more directly than the Dirichlet energy, which focuses on the flow field without accounting for the actual sampling process.

\subsection{Efficiency of Time-Splitting Estimation}
\label{subsec:boundary_efficiency}

A potential practical concern regarding \method\ is the reliance on a pre-trained auxiliary model to compute the cumulative modeling complexity across the generative timeline. However, we demonstrate that deriving the time partitioning with our approach incurs negligible computational overhead in practice. To validate this, we progressively estimate the complexity using increasingly lightweight network configurations:
\begin{enumerate}
    \item \textbf{Full SiT-XL/2:} A massive, fully trained baseline.
    \item \textbf{Full SiT-S/2:} A significantly smaller architecture, fully trained.
    \item \textbf{SiT-S/2 (50K Iterations):} The same small architecture, but trained for only a fraction of the standard training schedule (50K versus the standard 400K iterations).
    \item \textbf{SiT-S/2 (10\% Data):} The same small architecture, trained on only 10\% of the ImageNet dataset for a fraction of the standard training steps.
\end{enumerate}

Remarkably, all four configurations yield nearly identical complexity curves and produce the exact same temporal boundary placements. This demonstrates that the flow dynamics captured by our complexity measures are fundamentally robust to architectural scale and training duration, particularly when the objective is to assess a small number of parameters ($N$).
\section{Related Work}
\label{sec:related_work}

\paragraph{Temporal Specialization in Diffusion Models.}
Recognizing that global networks struggle to efficiently model heterogeneous generative trajectories, several works explore temporal specialization. Cascaded models~\citep{ho2022cascaded} divide generation across independent networks, but partition by spatial resolution rather than time. Operating explicitly on the time axis, eDiff-I~\citep{balaji2022ediff} and MEME~\citep{lee2024multi} train expert denoisers for specific noise intervals, yet identifying optimal transition boundaries requires exhaustively expensive empirical searches. Other approaches use time-conditioned Mixture-of-Experts (MoE) and dynamic task routing to bypass strict boundaries. Models like Denoising Task Routing (DTR)~\citep{park2023denoising}, Switch Diffusion Transformers~\citep{park2024switch}, RAPHAEL~\citep{xue2023raphael}, and recent MoE-transformers~\citep{cheng2025diff} dynamically allocate compute based on the active timestep. While effective, these learned, black-box routing mechanisms are notoriously difficult to stabilize, prone to routing collapse, and lack guarantees for balanced representational work. In contrast, our method provides a mathematically principled, search-free algorithm to optimally partition the timeline.

\paragraph{Approximation Theory in Neural Networks.}
Approximation theory provides rigorous bounds on the representational capacity required to fit complex functions. Classic theorems by Barron~\citep{barron2002universal} bound feedforward network errors using the target function's spectral complexity, formally linking high-frequency spatial fluctuations to larger required parameter counts. For deep architectures, Yarotsky~\citep{YAROTSKY2017103} extended these bounds to Sobolev spaces, defining error decay rates governed by maximal high-order derivatives along continuous curves. Beyond static functions, continuous vector field representation has been analyzed in Neural ODEs, bounding trajectory complexity and integration error via Jacobian traces and Hutchinson estimators~\citep{finlay2020train, kelly2020learning, hutchinson1990stochastic}. While profound, applying these frameworks to generative models is typically hindered by the intractability of evaluating high-dimensional spectral norms or exact high-order derivatives. Our work bridges this gap by translating abstract theoretical bounds into tractable monitor functions. By specifically bounding spectral complexity via Dirichlet energy and trajectory error via path acceleration, we turn classical approximation theory into a practical tool for allocating network capacity over time.

\paragraph{Scaling Up Generative Models.}
The predictable improvement of neural networks with increased capacity, formalized as scaling laws~\citep{kaplan2020scaling, hoffmann2022training, liang2024scaling}, remains foundational in deep learning. In visual generative modeling, this principle has driven a push toward larger architectures. Breakthroughs like the Diffusion Transformer (DiT)~\citep{peebles2023scalable}, Scalable Interpolant Transformers (SiT)~\citep{ma2024sit}, and modern large-scale models~\citep{flux2024, wu2025qwen} demonstrate that aggressively scaling parameters yields predictable improvements in sample fidelity. However, standard continuous-time frameworks deploy monolithic architectures, applying this massive parameter budget at every integration step. Consequently, adhering to scaling laws incurs a proportional and often prohibitive increase in inference-time computational cost (FLOPs). Our approach decouples parameter scaling from inference cost: by distributing the expanded capacity across the temporal axis, we allow total model capacity to scale according to these laws while ensuring the active parameter count at any given timestep remains constant.

\section{Conclusion}
\label{sec:conclusion}

We introduced Complexity-Balanced Splitting (\method), a principled framework for temporal capacity allocation in continuous-time generative models. To overcome the inefficiency of scaling monolithic architectures, we framed timeline partitioning as a domain decomposition problem in the context of approximation theory. Leveraging de Boor's equidistribution principle, we demonstrated that generative performance is maximized by dividing the timeline into segments of equal representational burden, rigorously quantified using either spatial Dirichlet energy or temporal path acceleration. While the latter proved to be slightly advantageous, by presenting and evaluating both, we cover the two central approaches to monitor the complexity of a flow field in terms of its analytical regularity.

Empirical evaluations across diverse architectures (SiT, JiT, UNet) confirm that \method\ significantly improves synthesis quality without increasing per-step inference costs. Furthermore, because these complexity metrics reflect fundamental geometric properties of the generative trajectory, optimal boundaries can be estimated pre-training with negligible overhead, entirely eliminating the need for costly empirical searches.

This work focuses on the temporal axis and achieves close to optimal solutions in this space. However, we believe this playground should be expanded, and the equidistribution principle should be used to derive other forms of splitting in new domains. For instance, spatial splitting, in which individual tokens are routed between networks, represents a potentially stronger direction. Deriving the proper monitor functions for this is expected to be a challenge, and we leave this topic for future work.

Ultimately, \method\ provides a mathematically grounded, search-free solution to decouple total parameter capacity from inference costs, enabling massively scaled models that focus compute exactly where the generative dynamics demand it most.

\bibliography{references}
\bibliographystyle{abbrv}

\clearpage
\appendix

\section{Bounding Spectral Complexity via Dirichlet Energy}
\label{sec:appendix_a}

In this section, we formally derive the relationship between the spectral complexity $C_{v_t}$ and the Dirichlet energy $E_D(v_t)$ using the Cauchy-Schwarz inequality. 

Recall the definition of the spectral complexity for the vector field $v_t$ is given by
\begin{equation}
    C_{v_t} = \int_{\mathbb{R}^d} \|\omega\| \|\hat{v}_t(\omega)\| d\omega.
\end{equation}

To bound this integral using the $L^2$ norm of the gradient (which corresponds to the Dirichlet energy), we must account for the fact that the $L^1$ norm of a function over $\mathbb{R}^d$ cannot be bounded by its $L^2$ norm without an additional decaying weight or an assumption of bounded support. In practical physical and neural network applications, we can safely assume the flow field is effectively band-limited. That is, its spectral energy is negligible outside a frequency ball of radius $\Omega_{\text{max}}$, allowing us to restrict our domain of integration to $B(0, \Omega_{\text{max}})$.

The Cauchy-Schwarz inequality for integrals states:
\begin{equation}
    \left( \int_{D} f(\omega) g(\omega) d\omega \right)^2 \le \left( \int_{D} f(\omega)^2 d\omega \right) \left( \int_{D} g(\omega)^2 d\omega \right).
\end{equation}

We set $f(\omega) = 1$ and $g(\omega) = \|\omega\| \|\hat{v}_t(\omega)\|$ over the domain $D = B(0, \Omega_{\text{max}})$. Applying the inequality yields
\begin{equation}
    C_{v_t}^2 = \left( \int_{B(0, \Omega_{\text{max}})} 1 \cdot \|\omega\| \|\hat{v}_t(\omega)\| d\omega \right)^2 \le \left( \int_{B(0, \Omega_{\text{max}})} 1 d\omega \right) \left( \int_{B(0, \Omega_{\text{max}})} \|\omega\|^2 \|\hat{v}_t(\omega)\|^2 d\omega \right).
\end{equation}

The first term on the right-hand side is simply the volume of the $d$-dimensional ball of radius $\Omega_{\text{max}}$, which we denote as $V(\Omega_{\text{max}})$. Assuming the energy outside this ball is negligible, the second term can be extended back to $\mathbb{R}^d$ and related to the Dirichlet energy via Parseval's identity.

Recall Parseval's identity for the gradient of a function is given by
\begin{equation}
    \int_{\mathbb{R}^d} \|\nabla_x v_t(x)\|^2 dx = (2\pi)^d \int_{\mathbb{R}^d} \|\omega\|^2 \|\hat{v}_t(\omega)\|^2 d\omega.
\end{equation}

Substituting the definition of the Dirichlet energy, $E_D(v_t) = \frac{1}{2} \int_{\mathbb{R}^d} \|\nabla_x v_t(x)\|^2 dx$, we have:
\begin{equation}
    \int_{\mathbb{R}^d} \|\omega\|^2 \|\hat{v}_t(\omega)\|^2 d\omega = \frac{2}{(2\pi)^d} E_D(v_t).
\end{equation}

Plugging this equivalence back into our Cauchy-Schwarz bound, we arrive at the final inequality,
\begin{equation}
    C_{v_t}^2 \le V(\Omega_{\text{max}}) \frac{2}{(2\pi)^d} E_D(v_t).
\end{equation}

Defining the constant $K = \frac{2 V(\Omega_{\text{max}})}{(2\pi)^d}$, we obtain the relation used in the main text,
\begin{equation}
    C_{v_t}^2 \le K \cdot E_D(v_t).
\end{equation}

\begin{figure}[h]
    \centering
    \begin{subfigure}[b]{0.65\textwidth} 
        \centering
        \includegraphics[width=\textwidth]{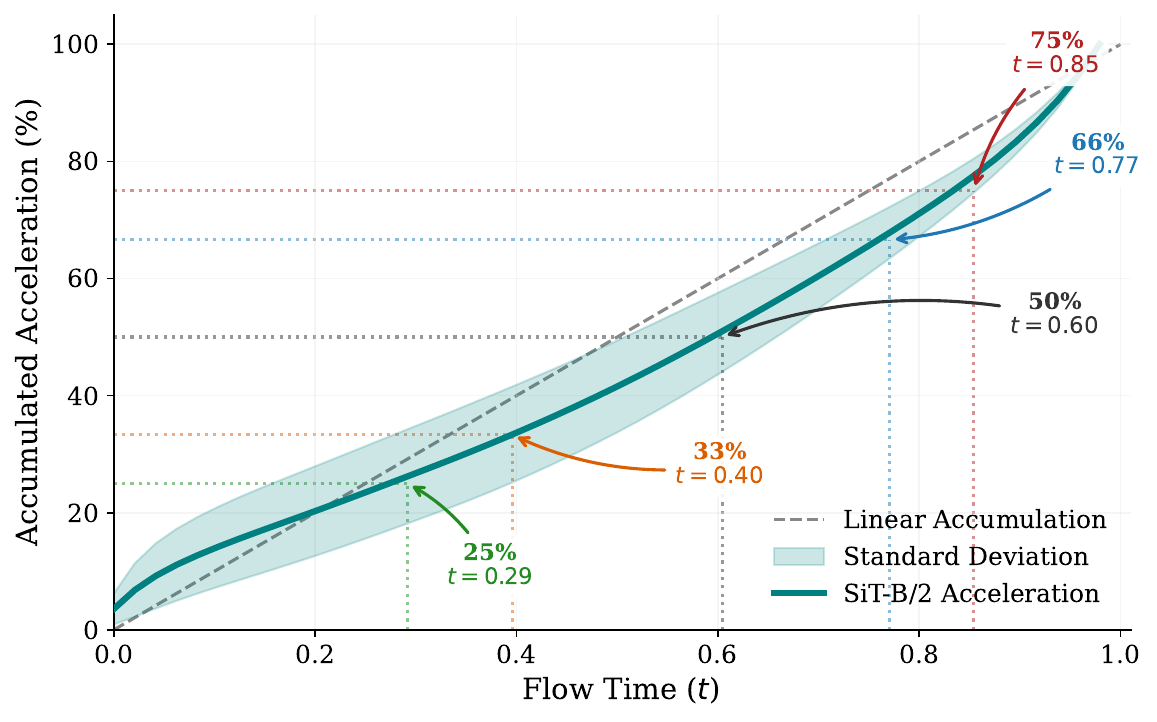}
        \caption{SiT (ImageNet-256)}
        \label{fig:accel_sit}
    \end{subfigure}
    
    \vspace{1em} 
    
    \begin{subfigure}[b]{0.65\textwidth}
        \centering
        \includegraphics[width=\textwidth]{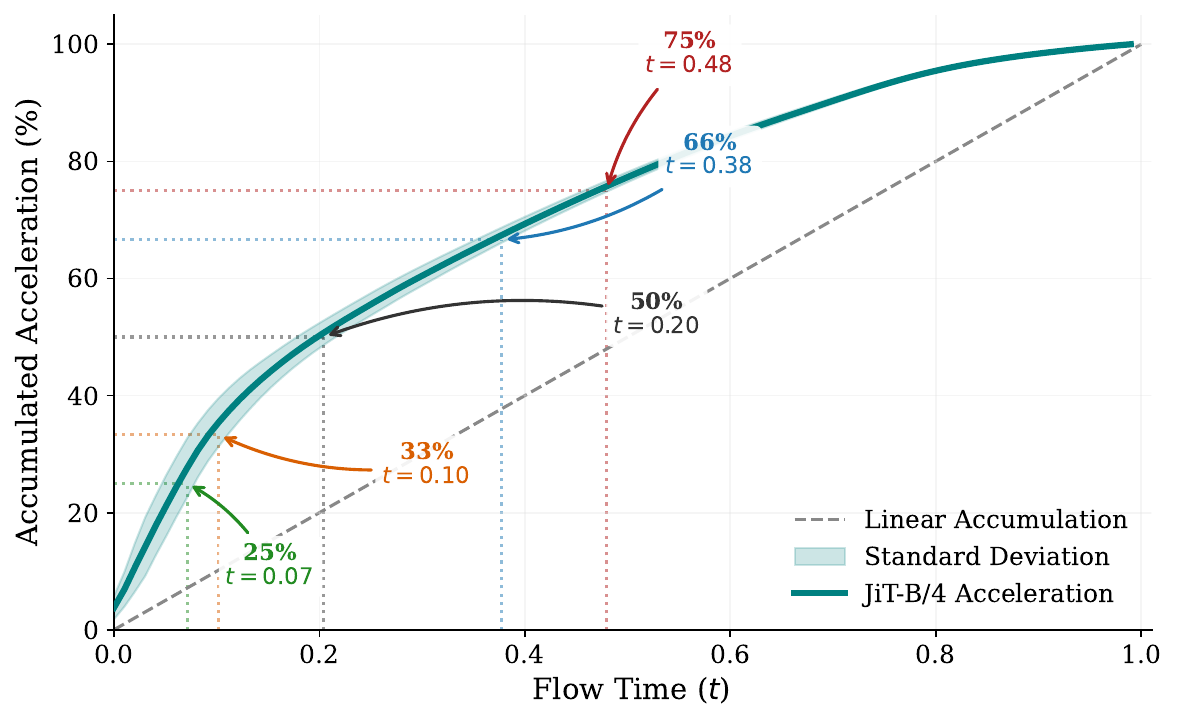}
        \caption{JiT (ImageNet-64)}
        \label{fig:accel_jit}
    \end{subfigure}
    
    \vspace{1em} 
    
    \begin{subfigure}[b]{0.65\textwidth}
        \centering
        \includegraphics[width=\textwidth]{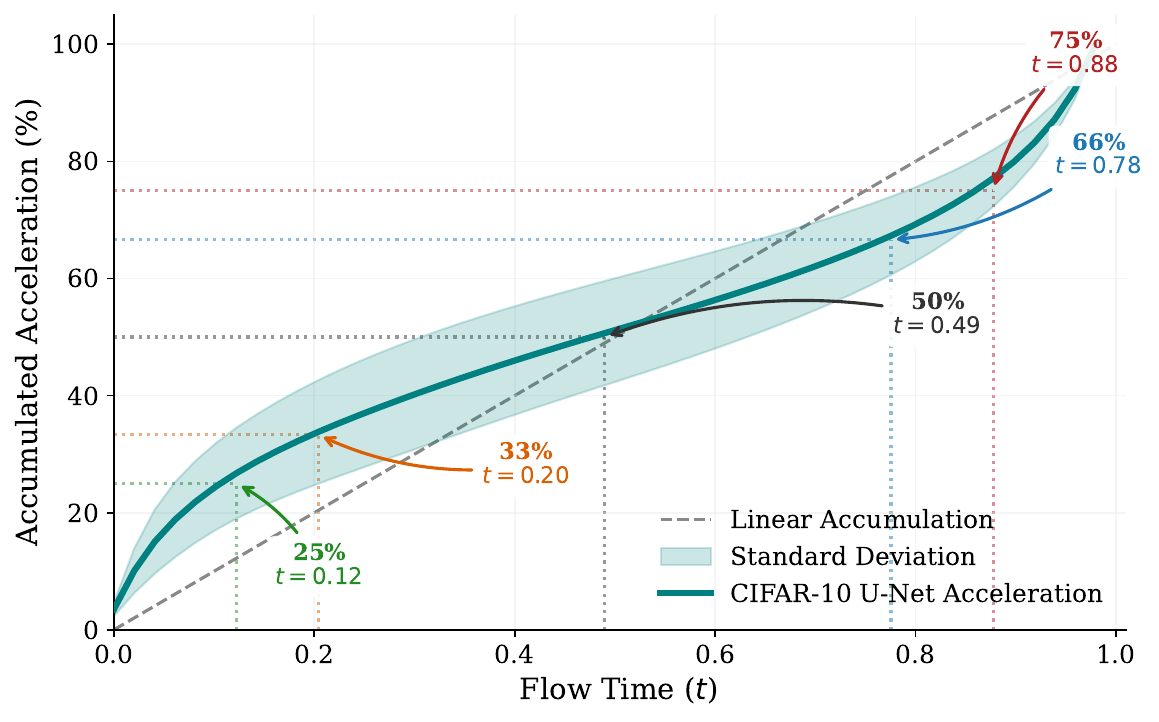}
        \caption{UNet (CIFAR-10)}
        \label{fig:accel_cifar}
    \end{subfigure}
    
\caption{
    \textbf{Cumulative path acceleration across datasets and models.} 
    We visualize the cumulative second-order time derivatives (path acceleration) over the generative timeline $t \in [0, 1]$ for SiT, JiT, and UNet.
}
    \label{fig:supp_cumulative_accel}
\end{figure}
\section{Cumulative Path Acceleration Analysis}
\label{sec:appendix_acceleration}

To further illustrate the temporal dynamics of the generative process and support the empirical findings in Section~\ref{sec:experiments}, Figure~\ref{fig:supp_cumulative_accel} visualizes the cumulative path acceleration (derived in Section~\ref{subsec:curvature_work}) across our three evaluated baselines: SiT on latent ImageNet-256, JiT on pixel-space ImageNet-64, and UNet on CIFAR-10.

As observed in the plots, the accumulation of geometric trajectory acceleration is distinctly non-linear across all architectures and data modalities.
By defining our temporal boundaries at intervals of equal accumulated acceleration, \method\ naturally adapts to these dataset-specific dynamics. It assigns narrower time intervals (and thus, higher localized parameter capacity) to the steepest phases of the curve. This visually confirms why our mathematically derived splits consistently avoid the capacity bottlenecks that degrade the performance of uniformly partitioned baselines.

\section{Implementation Details}
\label{sec:appendix_impl}

\paragraph{Hardware and Compute Resources.}
All experiments were conducted on a high-performance computing cluster equipped with NVIDIA H200 GPUs. To manage multi-GPU synchronization and efficiently scale our training workloads across the different architectures (SiT, JiT, and UNet), we utilized PyTorch DistributedDataParallel (DDP). Our underlying software environment was built on standard PyTorch and CUDA releases.

\paragraph{Hyperparameters and Configuration.}
The complete set of training, optimization, and sampling hyperparameters used across all experimental baselines and our specialized sub-networks is detailed in Table~\ref{tab:supp_hyperparams}. To ensure a rigorous and fair comparison, all models within a specific generative domain were trained and evaluated using these identical base configurations unless explicitly stated otherwise.

\begin{table}[h]
\centering
\caption{Training and sampling hyperparameters for the baseline architectures evaluated in our experimental setup.}
\vspace{0.5em}
\small
\setlength{\tabcolsep}{6pt}
\begin{tabular}{l c c c}
\toprule
\textbf{Parameter} & \textbf{SiT (ImageNet-256)} & \textbf{JiT (ImageNet-64)} & \textbf{UNet (CIFAR-10)} \\
\midrule
\rowcolor[gray]{0.95} \multicolumn{4}{l}{\textit{Architecture \& Data}} \\
Resolution & $256 \times 256$ & $64 \times 64$ & $32 \times 32$ \\
Space & Latent (VAE) & Pixel & Pixel \\
Patch Size & 2 & 4 & N/A \\
Batch Size & 256 & 1024 & 128 \\
\midrule
\rowcolor[gray]{0.95} \multicolumn{4}{l}{\textit{Training \& Optimization}} \\
Duration & 400K Iterations & 600 Epochs & 400K Iterations \\
Optimizer & AdamW & AdamW & AdamW \\
Learning Rate & $1 \times 10^{-4}$ & $1 \times 10^{-4}$ & $2 \times 10^{-4}$ \\
LR Schedule & Constant & Linear Warmup & Constant \\
Weight Decay & 0.0 & 0.0 & 0.0 \\
EMA Rate & 0.9999 & 0.9999 & 0.9999 \\
\midrule
\rowcolor[gray]{0.95} \multicolumn{4}{l}{\textit{Sampling}} \\
Method & Euler & Euler & Euler \\
Sampling Steps & 250 & 250 & 250 \\
CFG Scale & 1.5 & 2.9 & N/A \\
\bottomrule
\end{tabular}
\label{tab:supp_hyperparams}
\end{table}

\clearpage

\end{document}